\documentclass{article}

\usepackage[final,nonatbib]{tackling_climate_workshop_style}

\usepackage[utf8]{inputenc} 
\usepackage[T1]{fontenc}    
\usepackage{hyperref}       
\usepackage{url}            
\usepackage{booktabs}       
\usepackage{amsfonts}       
\usepackage{nicefrac}       
\usepackage{microtype}      
\usepackage[symbol]{footmisc}
\usepackage{multirow}
\usepackage{booktabs}
\usepackage{tikz}
\usetikzlibrary{arrows.meta, fit, calc, positioning, quotes,
                shadows, shapes.geometric, shapes.misc}

\tikzset{FlowChart/.style =
{
    box/.style = {rectangle, draw, fill=white,
               text width=##1, minimum height=8mm, align=center,
               inner sep=2mm, outer sep=0mm,
               },
    box/.default = 22mm,
    cbox/.style = {cross out=1m, draw, very thick,
                inner sep=2mm, node contents={}},
    fbox/.style = {rectangle, draw, densely dashed, inner sep=2mm},
    LA/.style = {semithick, -Triangle}
}}

\usepackage{bm}

\title{Aboveground carbon biomass estimate with Physics-informed deep network}

\author{%
  Juan Nathaniel\thanks{Corresponding author (jn2808@columbia.edu)}\\
  Columbia University\\
  NY, USA\\
   \And
   Levente J. Klein \\
   IBM Research \\
   Yorktown Heights, NY, USA \\
   \AND
   Campbell D. Watson \\
   IBM Research \\
   Yorktown Heights, NY, USA \\
   \And
   Gabrielle Nyirjesy \\
   Columbia University\\
   NY, USA \\
   \And
   Conrad M. Albrecht \\
   German Aerospace Center\\
   Weßling, Germany \\
}

\begin{document}
\maketitle

\begin{abstract}
  The global carbon cycle is a key process to understand how our climate is changing. However, monitoring the dynamics is difficult because a high-resolution robust measurement of key state parameters including the aboveground carbon biomass (AGB) is required. 
  Here, we use deep neural network to generate a wall-to-wall map of AGB within the Continental USA (CONUS) with 30-meter spatial resolution for the year 2021. We combine radar and optical hyperspectral imagery, with a physical climate parameter of SIF-based GPP. Validation results show that a masked variation of UNet has the lowest validation RMSE of 37.93 ± 1.36 Mg C/ha, as compared to 
  52.30 ± 0.03 Mg C/ha for random forest algorithm. Furthermore, models that learn from SIF-based GPP in addition to radar and optical imagery reduce validation RMSE by almost $10\%$ and the standard deviation by $40\%$. Finally, we apply our model to measure losses in AGB from the recent 2021 Caldor wildfire in California, and validate our analysis with Sentinel-based burn index. 
\end{abstract}

\section{Introduction}
Aboveground carbon biomass (AGB) is an important component to monitor carbon cycle on the local \cite{albrecht2022monitoring, anaya2009aboveground} and global scale \cite{zhang2020fusion, duncanson2019importance}. 
A recent state-of-the-art light detection and ranging (LiDAR) mission from NASA's Global Ecosystem Dynamics Investigation (GEDI) generates a global yet non-continuous sparse measurements of vegetation parameters, including AGB estimates with a 60-meter along-track and 600-meter across-track gaps between footprints \cite{dubayah2021}. Here, we apply a Physics-informed deep network to generate a 30-meter resolution, wall-to-wall continuous AGB estimate in CONUS for the summertime (June-August) 2021 period. The model is trained on more than one million GEDI footprints using a combination of radar and optical imagery \cite{nico2022}. In addition, we incorporate a measure of photosynthetic intensity as captured by solar-induced fluorescence (SIF)-based gross primary production (GPP). SIF-based GPP is one of the key physical parameters regulating AGB \cite{li2019mapping, wang2020photochemical}. We validate our results using field-based AGB observations and evaluate their consistency across climate zones. Finally, we use our high resolution AGB map to assess the impact of wildfire, in terms of how much carbon biomass had been lost to the environment.

\section{Methodology}
\begin{figure}[!htb]
    \centering
    \begin{center}
        \begin{tikzpicture}[FlowChart, node distance = 0mm and 8mm]
            \node (p1a) [box] {Sentinel-1 data};
            \node (p1b) [box, right=4mm of p1a] {Bilinear resampling};
            \node (p2a) [box, below=4mm of p1a] {Sentinel-2 data};
            \node (p2b) [box, right=4mm of p2a] {Near-cloud-free filtering};
            \node (p3a) [box, below=4mm of p2a] {SIF-based GPP};
            \node (p3b) [box, right=4mm of p3a] {Summertime averaging};
            \node (p4a) [box, below=4mm of p3a] {GEDI/Field-based AGB};
            \node (p4b) [box, right=4mm of p4a] {Gridcell matching};
            \node (p5) [fbox, fit=(p1a)(p1b)(p2a)(p2b)(p3a)(p3b)(p4a)(p4b)]  {};
            \node[above right] at (p5.north west){\emph{Data processing}};
            
            \node (p6a) [box, right=8mm of p2b] {Training datacube};
            \node (p6b) [box, right=8mm of p3b] {Validation datacube};
            \node (p7) [fbox, fit=(p6a)(p6b)]  {};
            \node[above right] at (p7.north west){Datacube};
            
            \node (p8) [box, right=42mm of p1b] {Hyperparameter optimization};
            \node (p8a) [box, below=4mm of p8] {Training};
            \node (p8b) [box, below=4mm of p8a] {Trained model};
            \node (p8c) [box, below=4mm of p8b] {Validation};
            \node (p9) [fbox, fit=(p8)(p8a)(p8b)(p8c)]  {};
            \node[above right] at (p9.north west){\emph{Modeling}};
            
            
            \coordinate[below=0mm of p1a.east] (c1);
            \coordinate[below=0mm of p2a.east] (c2);
            \coordinate[below=0mm of p3a.east] (c3);
            \coordinate[below=0mm of p4a.east] (c4);
            \coordinate[below=0mm of p5.east] (c5);
            \coordinate[below=0mm of p6a.east] (c6);
            \coordinate[below=0mm of p6b.east] (c7);
            \coordinate[below=0mm of p8.south] (c8);
            \coordinate[below=0mm of p8a.south] (c9);
            \coordinate[above=0mm of p8c.north] (c10);
            
            \draw [LA]  (c1) to (c1 -| p1b.west);
            \draw [LA]  (c2) to (c2 -| p2b.west);
            \draw [LA]  (c3) to (c3 -| p3b.west);
            \draw [LA]  (c4) to (c4 -| p4b.west);
            \draw [LA]  (c5) to (c5 -| p7.west);
            \draw [LA]  (c6) to (p8.west);
            \draw [LA]  (c7) to (p8c.west);
            \draw [LA]  (c8) to (p8a.north);
            \draw [LA]  (c9) to (p8b.north);
            \draw [LA]  (c10) to (p8b.south);
        \end{tikzpicture}
    \end{center}
    \caption{Schematic overview that summarizes our data-processing steps at the leftmost panel and datacube generation during training and validation steps. The modeling process is iterated across different data setups (SIF/S1/S2, S1/S2, and S2-only) in an ablation study.}
    \label{fig:scheme}
\end{figure}
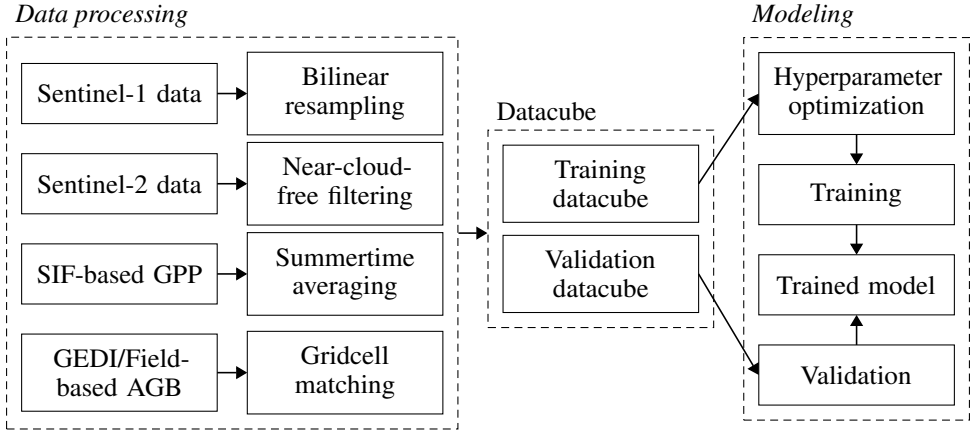

This section provides a detailed description of the data, models, and evaluation metrics used, and are summarized in Figure \ref{fig:scheme}.

\subsection{Data Processing}
GEDI is one of the current state-of-the-art spaceborne LiDAR missions to capture a detailed measure of ecosystem structure \cite{dubayah2020global}. However, the data generated is spatially sparse. Therefore, we attempt to produce a dense 30-meter resolution AGB estimate using multiple input features including radar and optical hyperspectral imagery from Sentinel-1 and Sentinel-2, as well as SIF-based GPP from the OCO-2 mission \cite{li2019mapping}. We use the vertical co-polarization (VV) and vertical/horizontal cross-polarization (VH) band of Sentinel-1, and the entire 12 bands of Sentinel-2. Furthermore, we produce near-cloud-free optical imagery by taking the median value of non-cloudy pixel as defined by the scene classification layer (SCL) of Sentinel-2. Finally, we perform grid cell geospatial matching to combine the dataset together. The validation sites are situated in the Northwestern America \cite{fekety2019annual} and New England \cite{tang2021lidar}. 






\subsection{Model and Experimental Setup}
\begin{figure}[h]
    \centering
    \includegraphics[width=0.7\linewidth]{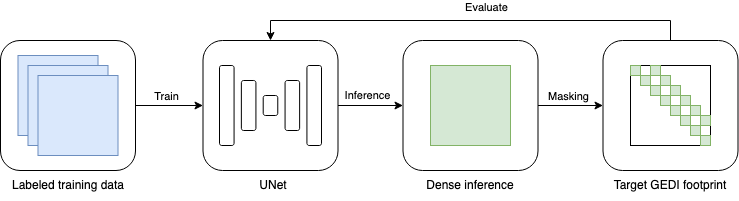}
    \caption{Our modeling workflow starting from training a multi-channel dataset using UNet and evaluating with sparse GEDI footprints}
    \label{fig:unet}
\end{figure}

We benchmark our deep network model against linear regressor (LR), random forests (RF) \cite{breiman2001random}, and extreme gradient boosting (XGBoost) \cite{chen2015xgboost} algorithms due to their reported robustness in climate-related tasks \cite{nathaniel2020bias}. Specifically, we implement a masked variation of UNet \cite{ronneberger2015u} due to the sparsity of our target AGB variable \cite{zhang2020map}. The models are optimized using a randomized grid search and a k-fold cross validation (k=5) approaches. We split the data into a 90-10 training-testing set, where set here refers to pixels for ML-based models and tiles for deep networks. We use Adam optimizer \cite{kingma2014adam} with a learning rate of 0.01 and root-mean-squared error (RMSE) as our loss function. For the deep network, we use a size 32 batching and implement a collection of image augmentations including random horizontal and vertical flipping, as well as cropping to a 512x512 image size. The full deep network modeling workflow is illustrated in Figure \ref{fig:unet}.

\section{Results and Discussion}
\begin{figure}[h]
    \centering
    \includegraphics[width=0.7\linewidth]{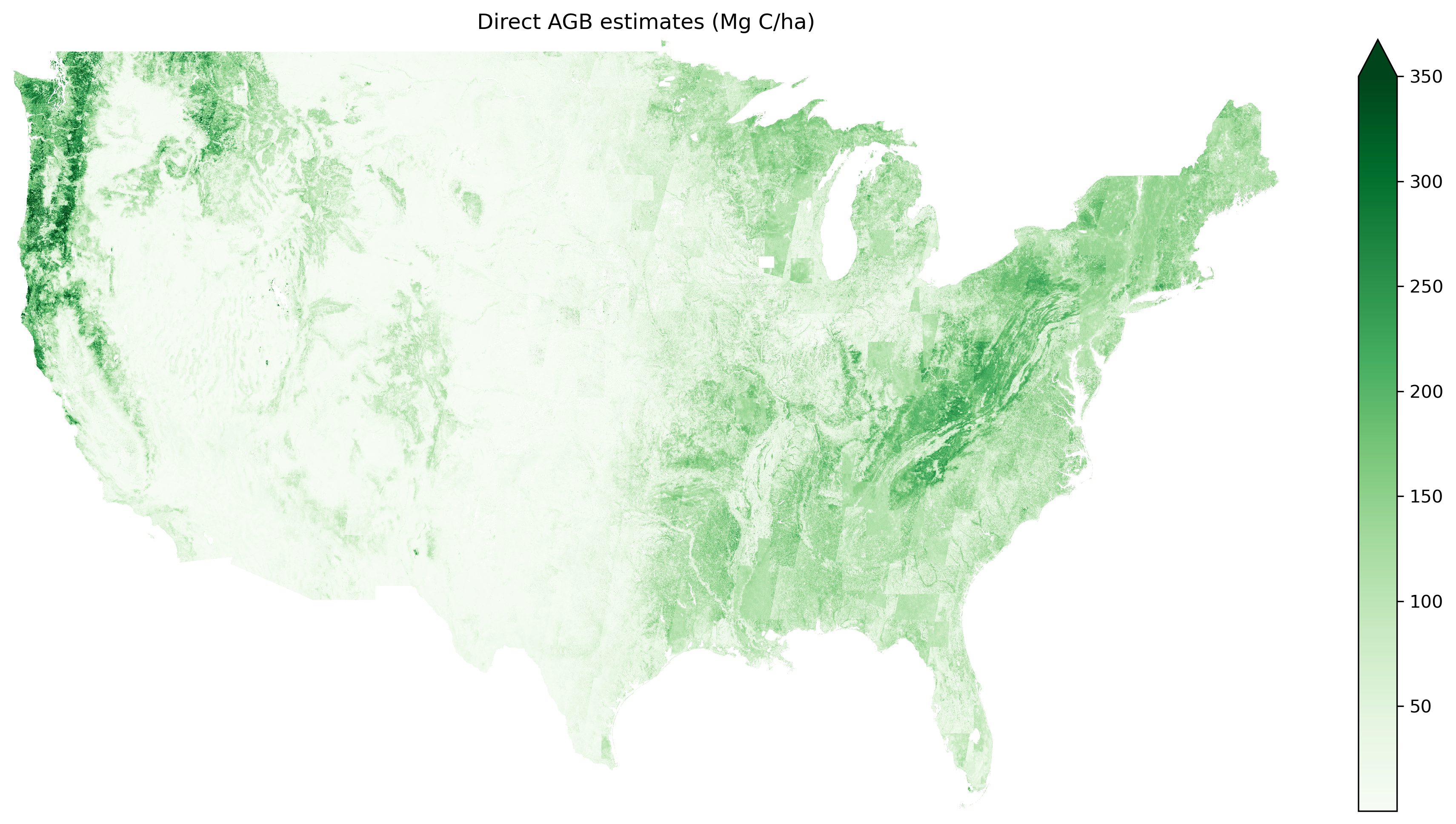}
    \caption{A 30-meter resolution AGB dense estimate for CONUS in the year 2021 summertime with Physics-informed masked variation of UNet}
    \label{fig:agb}
\end{figure}

This section summarizes and validates our results, showcasing the application of our AGB 30-meter resolution (Figure \ref{fig:agb}) to assess how much carbon has been released to the environment from a major wildfire event.

\subsection{Model Performance and Ablation Study}
As summarized in Table \ref{tab:result}, UNet has the lowest validation RMSE of 37.93 ± 1.36 Mg C/ha, as compared to 81.95 ± 0.01 Mg C/ha, 53.37 ± 0.05 Mg C/ha, and 52.30 ± 0.03 Mg C/ha for linear regressor, gradient boosting, and random forest algorithms respectively. Furthermore, models that learn from SIF-based GPP, in addition to radar and optical hyperspectral imagery have lower validation RMSE of 37.93 ± 1.36 Mg C/ha than 41.99 ± 3.23 Mg C/ha in the latter case. This error accounts for 20-40\% of the average AGB in CONUS, which is estimated by \cite{hu2016mapping, dubayah2021, santoro2021global} to be around 100 and 200 MgC/ha. Nonetheless, UNet still exhibits larger uncertainty across model runs than the other ML-based models. 

\begin{table}
    \centering
    \caption{Evaluation RMSE (Mg C/ha) for different combinations of inputs and models}
    \begin{tabular}[t]{lccc}\toprule
    Model & Inputs & Testing & Validation   \\ \midrule
    \hline
    \multirow{3}{*}{Linear Regressor} & SIF/S1/S2 & $66.07 \pm 0.06$ & $81.95 \pm 0.01$  \\
     & S1/S2 & $66.46 \pm 0.10$ & $84.33 \pm 0.00$ \\
     & S2-only & $67.10 \pm 0.11$ & $90.99 \pm 0.03$  \\
     \hline
    \multirow{3}{*}{XGBoost} & SIF/S1/S2 & $56.66 \pm 0.06$ & $53.37 \pm 0.05$  \\
     & S1/S2 & $57.35 \pm 0.05$ & $54.74 \pm 0.03$ \\
     & S2-only & $57.82 \pm 0.02$ & $54.81 \pm 0.26$  \\
     \hline
    \multirow{3}{*}{RF} & SIF/S1/S2 & $57.16 \pm 0.05$ & $52.30 \pm 0.03$  \\
     & S1/S2 & $58.05 \pm 0.03$ & $54.72 \pm 0.06$ \\
     & S2-only & $58.12 \pm 0.02$ & $54.88 \pm 0.18$  \\
     \hline
    \multirow{3}{*}{UNet} & SIF/S1/S2 & $\boldsymbol{48.83 \pm 0.19}$ & $\boldsymbol{37.93 \pm 1.36}$ \\
     & S1/S2 & $49.30 \pm 0.18$ & $41.99 \pm 3.23$ \\
     & S2-only & $50.35 \pm 0.43$ & $45.93 \pm 2.25$  \\
    \bottomrule
    \end{tabular}
    \label{tab:result}
\end{table}



\subsection{Consistency Check Across Climate Zones}

\begin{figure}[htb]
    \centering
    \includegraphics[width=0.5\linewidth]{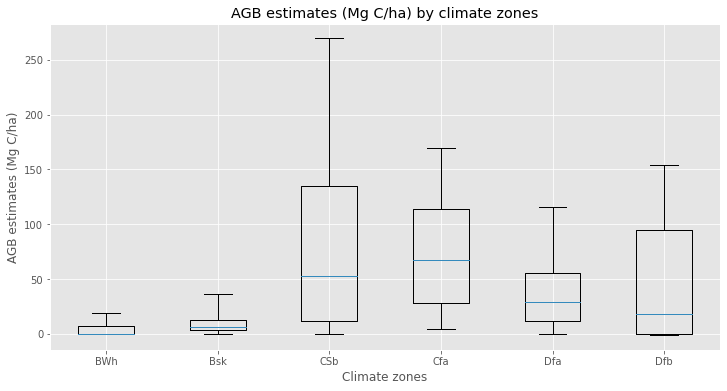}
    \caption{AGB estimates across different climate zones in CONUS - \textit{BWh}: arid, \textit{Bsk}: semi-arid, \textit{Csb}: dry summer temperate, \textit{Cfa}: humid subtropical, \textit{Dfa}: humid continental, \textit{Dfb}: humid continental}
    \label{fig:climate}
\end{figure}

Finally, we perform a consistency check to ensure that our AGB estimate agrees with expectations from literature. Figure \ref{fig:climate} illustrates the distribution of AGB estimate at pixel-level across the top-6 (by area size) Köppen-based climate zones in CONUS \cite{beck2018present}. As expected, the highest AGB esimates are observed at the dry summer temperate and humid subtropical regions ($\sim$ 70-300 MgC/ha), while the lowest ones are at the arid and semi-arid regions ($\sim$ 5-15 MgC/ha) \cite{harris2021global, yang2009aboveground, chave2014improved, santoro2021global}. 

\subsection{Application: Wildfire Impact Assessment}
\begin{figure}[htb]
    \centering
    \includegraphics[width=0.6\linewidth]{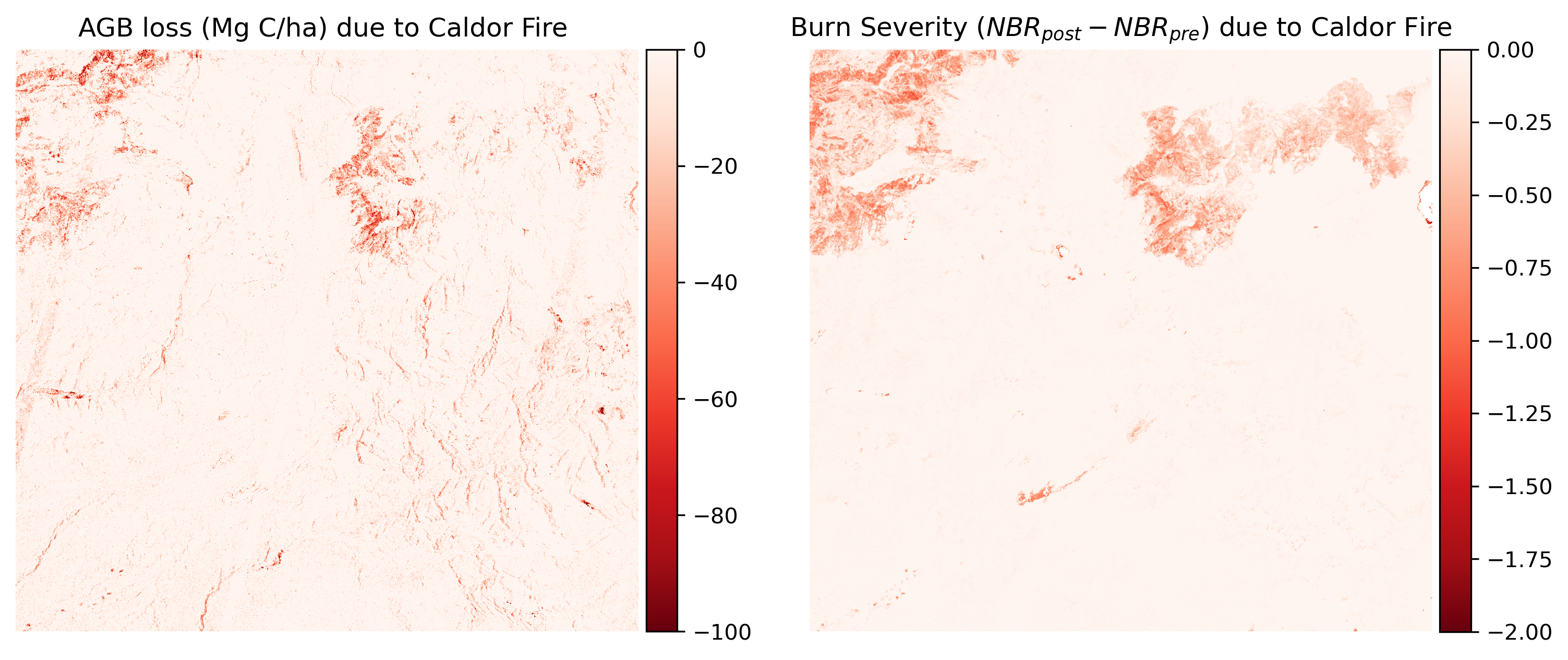}
    \caption{Left panel: AGB loss due to Caldor wildfire; Right panel: NBR index where lower value indicates more burning}
    \label{fig:fire}
\end{figure}

Lastly, we apply our AGB estimate to evaluate how much AGB has been lost after a major wildfire event \cite{zhou2020monitoring}. We analyzed the Caldor wildfire event at California in the year 2021. The left panel in Figure \ref{fig:fire} highlights the difference between AGB after (June 2022) and AGB before (June 2021) the fire, while the right panel indicates a normalized burn ratio (NBR) as derived from Sentinel-2 bands: $(\frac{(B08 - B12)}{(B08 + B12)})$; lower NBR value suggests a burned area. We observe a close relationship between impact (AGB loss) and intensity (NBR). A robust AGB loss estimate, therefore, could help institutions assess fire risk mitigation strategies and forecast potential fire hazard \cite{rodrigues2022wildfire}, among many other. 
\section{Conclusion}
In conclusion, we have demonstrated the transformation of sparse GEDI measurements into continuous map using multimodal sensing of optical and radar satellites and how the addition of physical parameter helps improve performance across models. And the use of a deep network, specifically the masked variation of UNet, further improves performance. We showed that our AGB estimate agrees with previous literature in terms of its consistency across climate zones. Lastly, we showcased how our AGB estimate can be used to assess wildfire impact, among many other interesting yet critical applications including for the evaluation of carbon credits and conservation projects for primary forests. 

\bibliography{tackling_climate_workshop}

\end{document}